# INTERACTION DESIGN FOR EXPLAINABLE AI

## OZCHI 2018 WORKSHOP

**Tuesday December 4th 2018 in Room 80.11.06 of RMIT Building 80**

As artificial intelligence (AI) systems become increasingly complex and ubiquitous, these systems will be responsible for making decisions that directly affect individuals and society as a whole. Such decisions will need to be justified due to ethical concerns as well as trust, but achieving this has become difficult due to the `black-box' nature many AI models have adopted. Explainable AI (XAI) can potentially address this problem by explaining its actions, decisions and behaviours of the system to users. However, much research in XAI is done in a vacuum using only the researchers' intuition of what constitutes a `good' explanation while ignoring the interaction and the human aspect.

This workshop invites researchers in the HCI community and related fields to have a discourse about human-centred approaches to XAI rooted in interaction and to shed light and spark discussion on interaction design challenges in XAI.

## ORGANIZERS

| **Prashan Madumal** | **Dr Ronal Singh** | **Joshua Newn** | **Prof Frank Vetere** |
| --- | --- | --- | --- |
| PhD Candidate | Research Fellow | PhD Candidate | Director |
| AI and Autonomy Lab | AI and Autonomy Lab | Interaction Design Lab | Interaction Design Lab |
| School of Computing and Information Systems, University of Melbourne | School of Computing and Information Systems, University of Melbourne | School of Computing and Information Systems, University of Melbourne | School of Computing and Information Systems, University of Melbourne |
| 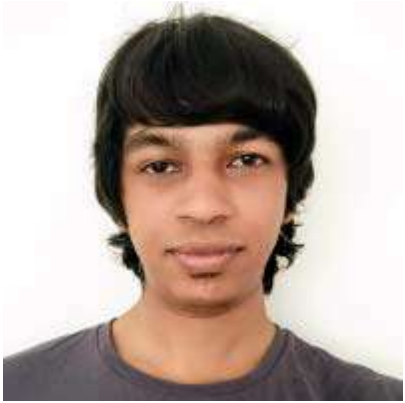 | 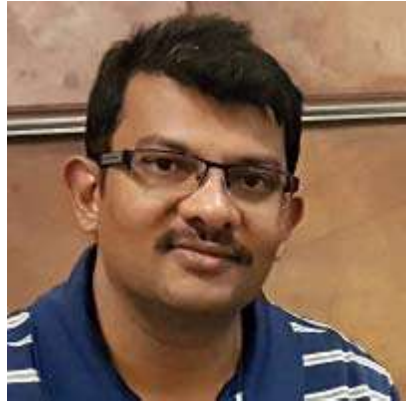 | 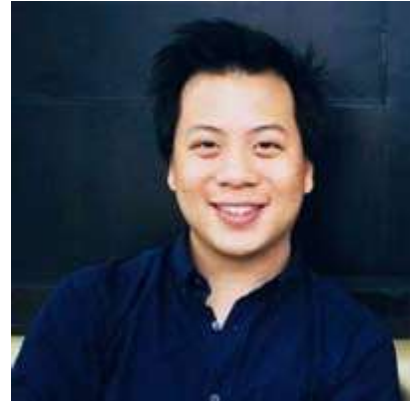 | 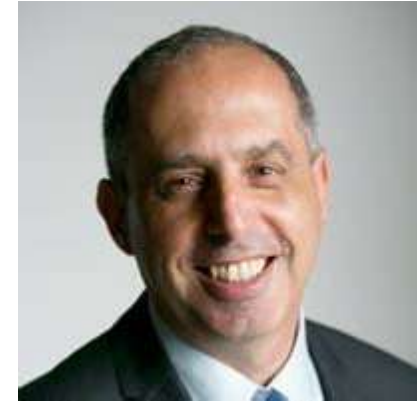 |

# ACCEPTED PAPERS

1. **But Why? Generating Narratives Using Provenance**, *Steven Wark, Marcin Nowina-Krowicki, Crisrael Lucero, Douglas Lange*

   - Steven Wark, Marcin Nowina-Krowicki, Defence Science & Technology Group, Edinburgh, SA 5111, Australia.

   - Crisrael Lucero, Douglas Lange, Space and Naval Warfare Systems Center Pacific, San Diego, CA 92110, USA.

2. **Demand-Driven Transparency For Monitoring Intelligent Agents,** *Mor Vered, Piers Howe, Tim Miller, Liz Sonenberg, Eduardo Velloso*

   - Mor Vered, Tim Miller, Liz Sonenberg, Eduardo Velloso, School of Computing and Information Systems, University of Melbourne, Australia.

   - Piers Howe, Melbourne School of Psychological Sciences, University of Melbourne, Australia.

3. **Transparency and Opacity in AI Systems: An Overview,** *Abdulrahman Baqais, Zubair Baig, Marthie Grobler*

   - Abdulrahman Baqais, Freelance Researcher, Dhahran, Saudi Arabia.

   - Zubair Baig, Marthie Grobler, CSIRO, Data61, Melbourne, Australia

4. **A Survey of Interpretable AI:** Aneesha Bakharia

   - Aneesha Bakharia, Institute of Teaching and Learning Innovation The University of Queensland

5. **Designing Explainable AI Interfaces through Interaction Design Techniques**, *Joshua Newn, Ronal Singh, Prashan Madumal, Eduardo Velloso, Frank Vetere*

   - Joshua Newn, Ronal Singh, Prashan Madumal, Eduardo Velloso, Frank Vetere, Microsoft Research Centre for Social Natural User Interfaces, School of Computing and Information Systems, University of Melbourne, Australia.

6. **A Grounded dialog model for Explainable Artificial Intelligence**, *Prashan Madumal, Tim Miller, Frank Vetere and Liz Sonenberg*
   - Prashan Madumal, Eduardo Velloso, Frank Vetere, Microsoft Research Centre for Social Natural User Interfaces, School of Computing and Information Systems, University of Melbourne, Australia.

   - Tim Miller, School of Computing and Information Systems, University of Melbourne, Australia.

# INVITED TALK
## Associate Professor Tim Miller
### Title: Explanation in Artificial Intelligence: Insights from the Social Sciences

**Abstract**

In his seminal book 'The Inmates are Running the Asylum: Why High-Tech Products Drive Us Crazy And How To Restore The Sanity' [2004, Sams Indianapolis, IN, USA], Alan Cooper argues that a major reason why software is often poorly designed (from a user perspective) is that programmers are in charge of design decisions, rather than interaction designers. As a result, programmers design software for themselves, rather than for their target audience; a phenomenon he refers to as the 'inmates running the asylum'. I argue that the currently hot-topic of 'Explainable artificial intelligence' risks a similar fate. Explainable artificial intelligence is the study of techniques to help people understand why algorithms have made particular decisions, with the aim of increasing trust and transparency of systems employing these algorithms. While the re-emergence of explainable AI is positive, most of us as AI researchers are building explanatory agents for ourselves, rather than for the intended users. But explainable AI is more likely to succeed if researchers and practitioners understand, adopt, implement, and improve models from the vast and valuable bodies of research in philosophy, psychology, and cognitive science; and if evaluation of these models is focused more on people than on technology.

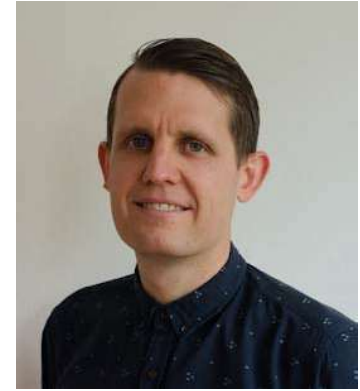

In this talk, I will demonstrate that most work in explainable AI is ignorant of the social sciences, argue why this is bad, and present several insights from the social sciences that are important for explanation in any subfield of artificial intelligence. The talk will be accessible to a general audience, and I hope it will be of particular interest to people working in artificial intelligence, social/cognitive science, or interaction design.

**Bio:**

Tim Miller an associate professor in the School of Computing and Information Systems at The University of Melbourne. His primary area of expertise is in artificial intelligence, with particular emphasise on:

- Human-AI interaction and collaboration
- Explainable Artificial Intelligence (XAI)
- Decision making in complex, multi-agent environments
- Reasoning about action and knowledge using automated planning

His research is at the intersection of artificial intelligence, interaction design, and cognitive science/psychology. His areas of education expertise is in artificial intelligence, software engineering, and technology innovation. He also have extensive experience developing novel and innovative solution with industry and defence collaborators and is a member of the AI and Autonomy Lab in the school.

# But Why? Generating Narratives Using Provenance

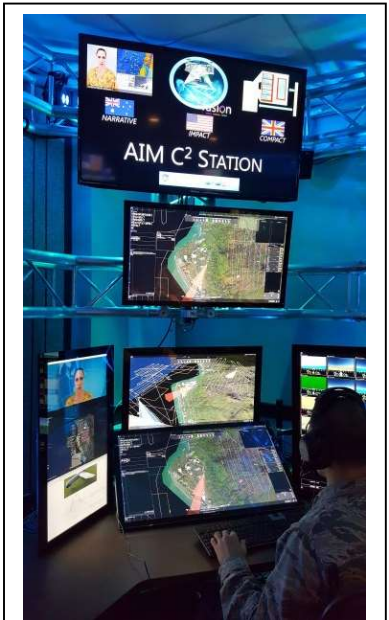

Figure 1: The AIM C2 system demonstrates how a single human operator can provide meaningful human control of over a dozen heterogeneous autonomous vehicles. AIM integrates several AI systems to support the human operator, and a provenance tracking system that allows narrative explanations to be generated of any changes in the system.

**Steven Wark**
**Marcin Nowina-Krowicki**
Defence Science & Technology Group
Edinburgh, SA 5111, Australia.
steven.wark@dst.defence.gov.au
marcin.nowina-krowicki@dst.defence.gov.au

**Crisrael Lucero**
**Douglas Lange**
Space and Naval Warfare Systems Center Pacific
San Diego, CA 92110, USA.
clucero@spawar.navy.mil
dlange@spawar.navy.mil

## Abstract
In human-machine partnerships, an AI system needs to not only explain its reasoning, but also establish the context that allows a human domain expert to correctly interpret the significance of this information. This paper describes how we have used the World Wide Web Consortium Provenance (W3C PROV) model to represent the flow of information in a Command & Control (C2) system for multiple autonomous vehicles. The PROV model allows us to decompose the C2 system into functional elements and identify the flow-on effects of any changes to the system inputs. A structured narrative is automatically generated to establish the context and explain to a human supervisor how the C2 system has produced its outputs, and how any changes may affect the tasking of one or more autonomous vehicles. This approach can be readily generalized to other domains wherever human-machine partnerships between AI and human domain experts are needed to handle large, complex and changing data.

## Author Keywords
Explainable AI; Data Analytics; Narrative Visualisation; Autonomy; C2 systems.

## ACM Classification Keywords
• **Human-centered computing** → Visualization
• **Human-centered computing** → Empirical studies in HCI

## Introduction
Human-machine partnerships are increasingly being used to analyse and make decisions with large, complex, and changing data. In these partnerships, a human domain expert relies on AI to manage and process the data, and generate outputs that helps them to understand its significance. Additionally, any changes to the data may have unexpected effects on the outputs generated. The system needs to not only provide an explanation of these changes, but to also set the context that allows the human to correctly interpret the explanation, especially for multiple changes or flow-on effects. Narrative, or storytelling, provides a mechanism to set and manage this context.

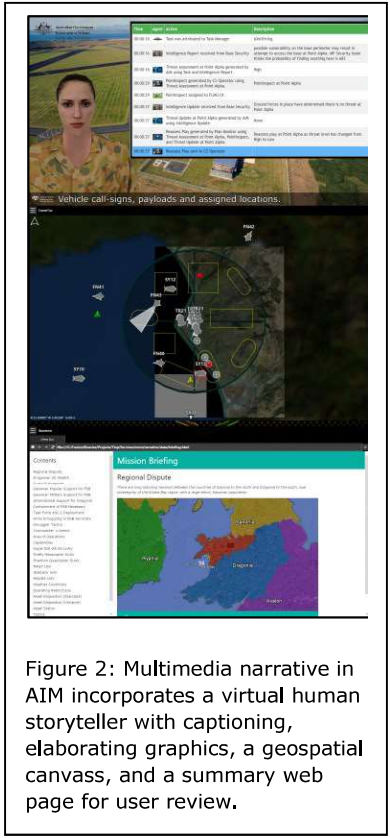

Figure 2: Multimedia narrative in AIM incorporates a virtual human storyteller with captioning, elaborating graphics, a geospatial canvass, and a summary web page for user review.

We have applied the W3C PROV model[1] to represent how information is fused and transformed within a C2 system. This allows us to trace what information and processing stages have contributed to any particular output, and thus generate a narrative explanation of how the output was obtained. Further, we can track the flow-on effects of any changes in the system and either explain the consequences of any changes, or explain why a particular output has changed.

### C2 of Autonomous Vehicles

Traditionally, a team of human operators have been responsible for remotely controlling a single unmanned vehicle. The Allied IMPACT (AIM) C2 system (see Figure 1) has been developed by an international team as part of The Technical Cooperation Program's (TTCP) Autonomy Strategic Challenge (ASC), to demonstrate how a single human operator could supervise a team of over a dozen heterogeneous autonomous vehicles [5]. AIM integrates various AI systems to support the human operator, including the autonomy on-board the vehicles. In a complex C2 system such as AIM, a change in a single piece of information can have unexpected flow-on effects on: the fused picture displayed; the operating constraints applied to the vehicles; what tasking options are available to the operator; and what the expected outcomes of these options may be. From these changes in their tasking, it is difficult for the operator to keep track of the effects, and understand what may have caused an unexpected change or response in the system.

A provenance tracking system has been developed in AIM to track the dependencies between the information flowing through the system as it is transformed and fused by the AI components. In order to understand the significance of any particular change, the operator may need to switch context and associated mental models as they deal with multiple vehicles and changes. To achieve this, multimedia storytelling using a virtual human storyteller is used to 'set the scene' and explain these changes, so that the operator can understand the context of an unfolding situation and make the appropriate decisions to achieve the mission goals (see Figure 2).

### Multimedia Narrative

Narrative was used in AIM as it is a common method employed by humans to effectively convey understanding of complex situations to each other. Narrative immerses the audience within a situation and allows them to understand the main events and actors, the important relationships between them, and what the consequences may be [2]. Multimedia narrative coordinates spoken or written narration, with other display modalities are used to rapidly convey key features and relationships. Different display modalities can be chosen to convey the appropriate level of trust in the information provided. Virtual human storytellers

---

[1] http://www.w3.org/TR/prov-dm

> **Rhetorical Structure Theory (RST)**
>
> In RST, one element known as the satellite may be related to a *nucleus* element by a rhetorical relation. A nucleus may have more than one satellite, and satellites may themselves have multiple satellites, forming a hierarchy describing how each part of a narrative is related to the other.
>
> Figure 3: RST satellite relation
>
> Our implementation of RST uses a simplified set of 5 satellite relationships that are also used to determine the sequencing of the content within the narrative:
>
> - Preparation
> - Background
> - Elaboration
> - Continuation
> - Conclusion

can use culturally appropriate behaviors to achieve effective engagement, manage trust, and convey uncertainty, risk, importance and urgency.

As with human interactions, narrative may not always be the most appropriate method of conveying information. The level of detail needed in the narrative depends on the information requirements of the target audience, based on their prior experience and interaction with the system. Ideally the narrative should reduce to traditional visualizations as the audience becomes contextually situated in the 'story'. This is achieved in AIM by maintaining a model of the operator's requirements, prior knowledge, and interactions with the narrative system.

The narrative system we implemented in AIM uses Rhetorical Structure Theory (RST) to provide a framework for achieving and maintaining coherence within a multimedia presentation (see Figure 3) [1]. Each element in the multimedia presentation has a rhetorical relationship to another element that describes its narrative role. A multimedia presentation structured in this way forms a graph, or more precisely a hierarchical tree, that links down to the central conceptual element of the presentation. Various pruning strategies can be applied to the graph to produce coherent narratives with different levels of detail that are tailored to meet the current information requirements of the operator [4].

## Provenance

The provenance of information in AIM is tracked and recorded by a component that is connected to a central messaging hub. It listens to the many different components in the system as they interact and enact change. The provenance information is recorded using a PROV java library, which implements the W3C Provenance Specification of the PROV Data Model. The PROV data model allows recording the relationships between *entities* (reports/assessments/tasks), *activities* (actions/events) and the *agents* (software components/vehicles/people) in the system, describing how things were created, changed or delivered. The provenance system uses PROV-Templates for declaring decision characterization and real-time instrumentation [3]. This PROV structure not only provides a framework for explaining how a particular report, assessment or task was obtained, but in turn allows us to track the corresponding flow-on effects of any particular change.

## Narrative Generation

The records captured in the PROV model in AIM are used by a narrative generation module to automatically construct a narrative explanation of the changes. For our initial implementation a two-stage approach was chosen to allow different options to be explored:

1. The PROV representation is mapped to an RST structure that represents different levels of detail in the generated narrative.

2. The RST structure is rendered as a multimedia explanation based on a model of the operator's inferred information needs, previous interactions with the AIM system and current workload. The RST structure can be collapsed and pruned in different ways to suit these requirements. Different rendering modalities can also be chosen based on available display options and user preferences.

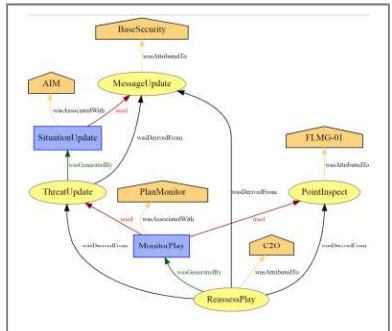

Figure 4: Simplified provenance model of processing flow in AIM for an intelligence report update.

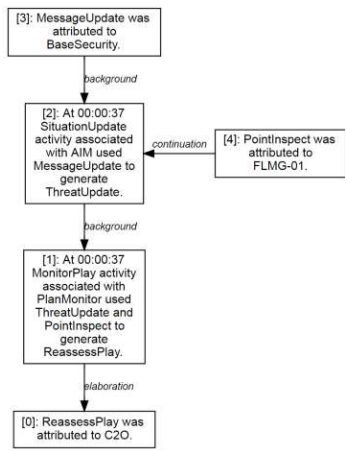

Figure 5: Narrative structure of intelligence update from perspective of the invoked reassessment task.

Our implementation in the AIM system was based around the use-case of tracking the flow-on effects of updates to an intelligence report. Reporting to the operator, the consequent need to reassess tasking for an air vehicle, and explaining to the operator why this tasking should be reassessed. This involves a number of agents, activities, and entities within AIM that are represented in the PROV model (see Figure 4).

Our approach generates narratives from the perspective of a particular agent or entity within the model, by recursively following the PROV relationships associated with this agent or entity and mapping this to an RST structure. The mapping chosen was designed to produce a hierarchy that summarized the derivations of entities at different levels in the generated hierarchy. Each level of the hierarchy provides progressively more information about the agents and entities that have influenced the targeted agent or entity. In this way, narratives can be generated for, for example, explanations of why the AIM system has suggested that tasking of this vehicle be reassessed (see Figure 5).

A number of different options are available for presenting these narratives to the AIM operator. Our initial implementation renders the timeline of activities in a table while the virtual human provides a voice-over describing the sequence of activities and highlighting any associated locations or physical actors on a geospatial display. Given the AIM operator may need to attend to multiple events concurrently, the system provides them with a notification when the provenance system has detected a change in the system state and a narrative is available for presentation.

## Conclusion

This paper demonstrates how the provenance tracking system can be used to automatically generate a structured explanation of an event in the AIM system. The approach used can be extended to provide explanations for arbitrarily complex situations and work-flows within AIM. Significantly, this approach allows tailored explanations to be dynamically generated without any prior knowledge of the dependencies within the AIM system, or the environment in which it is being deployed.

This approach can be readily generalized to other domains wherever the outputs of AI systems need to be explained to a human domain expert.

# Demand-Driven Transparency For Monitoring Intelligent Agents

Mor Vered, Piers Howe, Tim Miller, Liz Sonenberg, Eduardo Velloso
School of Computing and Information Systems
Melbourne School of Psychological Sciences
University of Melbourne, Australia
{mor.vered,pdhowe,tmiller,l.sonenberg,eduardo.velloso}@unimelb.edu.au

## Introduction

Designing *transparent* intelligent agents that can convey at least some information regarding their internal reasoning processes, is considered an effective method of increasing trust. How people *interact* with such transparency information to gain situation awareness while avoiding information overload is currently an under-explored topic.

We present a new mode of transparency acquisition that we call *Demand-Driven Transparency* (DDT). This model provides users with basic, course-grained control over which transparency information they acquire about specific components of the IA's reasoning process. In this way, the human operator can acquire information on demand. We contrast this approach with a baseline approach we refer to as *Sequential Transparency* (ST), in which the human operator must review the information regarding the IA's reasoning process in a predefined order.

We found that the DDT approach maintained the same performance when contrasted with the ST approach while reducing the workload. The DDT approach also managed to significantly increase the participants' trust in the IA while maintaining the same level of usability. These findings indicate that interactive explanation is a promising area for increasing transparency of intelligent agents.

## Intelligent Agent Transparency

Using Endsley's model of situation awareness [2] as a basis, we define a four-level model of IA transparency, which we call the *Endsley-based Transparency Model* (ETM). The four levels represent the reasoning process of the IA.

1. *No Knowledge* (Level 0): The IA's decision(s) with no corresponding explanation.

2. *Perceived Input* (Level 1): Basic factual information about the input, *as perceived by the IA*, is made transparent to the user.

3. *Input Reasoning* (Level 2): The immediate inferences the IA makes based on Level 1 information. Examples would be the possible ramifications of weather conditions on the availability of an unmanned aerial vehicle.

4. *Plan Projection* (Level 3): The IA predictions regarding future events and the uncertainties of the occurrences of these future events.

*Process Model*
We define two simple process models for instantiating this SA-based model: *Sequential Transparency* and *Demand-Driven Transparency*.

*Sequential Transparency*
We characterize *Sequential Transparency* as the manner of acquiring information about the IA's reasoning process in predefined, ordered steps. The manner or substance of the information conveyed is controlled externally and provided in an identical manner for each decision, irrelevant of the context or the person using it. This approach has the advantages of leveling the field and maintaining a unified level of knowledge among all users while also making sure the operator will be exposed to specific information that may influence the decision making process. However, because this approach targets the lowest common denominator, we hypothesize that it may be redundant or time consuming for some of the users in terms of the amount of information displayed.

*Demand-Driven Transparency*
DDT puts the user in control of the flow of knowledge-state information. DDT allows a human operator/observer to not only determine the order at which to request the information but also the type of information requested. This allows for a much more personalized form of interaction.

When dealing with human operators, we must consider that different users employ different processes when making decisions [3]. As such there would be significant advantages for allowing the operator to request information at their own discretion. In this manner, the system is more flexible and better suited to the individual needs of each different user, which we argue should not only reduce workload but also improve trust [1].

## Evaluation

We ran an empirical evaluation replicating a highly complex, military planning task. Participants were tasked with the surveillance of a set of off-shore and on-shore assets using a range of unmanned vehicles (UxVs), following standard protocols and responding to alerts. The scenario was designed to be complex enough that the participants would need to rely on the advice of an external intelligent planning agent. In each scenario, the intelligent agent recommends two possible plans, from which the operator must either select one as being the 'best' plan (according to the mission objectives), or indicate that neither is suitable. We ran two versions of this experiment, one in which the different levels of the ETM model were given *by demand* and another in which the information was supplied *sequentially*.

Participants for this experiment included 36 undergraduate and graduate students from the our University. No prior knowledge was required except proficiency in the English language. To facilitate making the decision, the human operator was able to access different levels of transparency by navigating through the different tabs in the top right region of the user interface (see Figure 1).

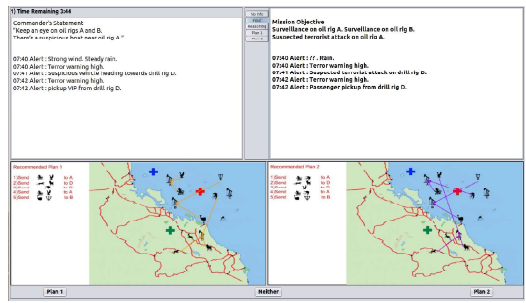

**Figure 1:** Experiment user interface.

*Results*
Our results confirmed that giving the operator control over which information they accessed led them to complete the tasks quicker, with an improvement in performance as seen in Tables 1 and 2. The results further showed that participants rated their trust in the system much higher and indicated the system as much easier to use.

We interpret these results as indicating that complex decision-making tasks may be better achieved with *interactive explanation*, which would allow people to pose specific questions and receive explanations. In these cases, users preferred looking at specific information to having to filter through to the information that they wanted. We hypothesise that interactive explanation that allows more fine-grained access to information and reasons would be more effective and perceived as more trustworthy.

|  | Performance(%) | | | Response Time(%) | | |
|---|---|---|---|---|---|---|
|  | M | SD | p-value | M | SD | p-value |
| DDT | 55.00 | 26.87 | 0.31 | **55.46** | 12.56 | **0.04** |
| ST | 50.56 | 21.24 | | 64.99 | 15.80 | |

**Table 1:** Comparison of performance and efficiency.

|  | Usability | | | Trust | | |
|---|---|---|---|---|---|---|
|  | M | SD | p-value | M | SD | p-value |
| DDT | **2.23** | 1.28 | 0.25 | **4.04** | 1.75 | **0.01** |
| ST | 2.13 | 1.25 | | 3.66 | 1.39 | |

**Table 2:** Perceptions of usability (higher is better) and trust (higher is better).

# Transparency and Opacity in AI Systems: An Overview

**Abdulrahman Baqais**
Freelance Researcher
Dhahran, Saudi Arabia
phd_abdulrahman@yahoo.com

**Zubair Baig**
CSIRO | Data61
Melbourne, Australia
zubair.baig@data61.csiro.au

**Marthie Grobler**
CSIRO | Data61
Melbourne, Australia
marthie.grobler@data61.csiro.au

## Abstract
Artificial Intelligence (AI) systems act as blackboxes, concealing their inner workings from the human operator. However, the level of opacity varies from one system category to another. We provide an overview of the level of opacity of these AI systems.

## Author Keywords
Opacity; transparency; human comprehension.

**ACM Classification Keywords**
H.5.m. Information interfaces and presentation (e.g., HCI): Miscellaneous; See http://acm.org/about/class/1998 for the full list of ACM classifiers. This section is required.

**Introduction**
Artificial Intelligence (AI) systems span a wide spectrum of applications. They can be perceived as a standalone system as found in robots, or as a major component of an autonomous vehicle or smart watch. AI systems typically adopt a blackbox approach, wherein the inner workings are concealed from the human interface, limiting the human operator's ability to fully comprehend the system's reasoning process in decision-making. We highlight the various categories of AI systems based on their levels of transparencies and opacity in terms of human interaction.

**Transparency vs Opacity**
Transparent systems typically provide software developers with visualizations of the system execution process. This facilitates a better understanding of the inner workings of the system, but often degrades system speed and performance [5]. AI systems differ from conventional software systems as they are more opaque in their execution, allowing software developers to focus on the software development process without concerning themselves with transparency enhancements. In addition, opacity ascertains that the system is hard to replicate; safeguarding their Intellectual Property and retaining their competitiveness in market.

Whilst AI systems are developed to improve process efficiencies in the industry, the current lack of transparency restricts clear understanding of the system's internal processes adopted for reaching a decision. This impacts liability and hinders preventative maintenance by human actors. Consider the application of AI by the judicial system. Such systems bear publicly-available data. The process of reaching a decision by an AI system is not always deterministic and as such it is almost impossible to challenge an outcome. As an example, an investigation conducted by ProPublica prompted the drafting and enactment of New York City's algorithmic accountability bill to confirm that the COMPAS (Correctional Offender Management Profiling for Alternative Sanctions) [4] risk assessments were more likely to erroneously identify black defendants as presenting a high risk for repeating offences, at almost twice the rate as white defendants (43% vs 23%). Due to the opaque nature of the AI system's risk assessment calculations, neither defendants nor the court systems utilizing COMPAS had visibility into why such assessments yielded significant rates of mislabeling based on race.

Whilst opacity is popularly adopted as a property in AI system design, incorporating an element of transparency in these systems would facilitate efficient testing and debugging, albeit with degraded AI system performance [5]. The movement to transparent AI is enforced through the introduction of the General Data Protection Regulation (GDPR) [2]. This regulation restricts AI systems that require large volumes of input data to train, from accessing data elements that hold privacy concerns. The GDPR further mandates that data-driven algorithms explain how the data has been processed to reach decisions [2]. Consequently, it is expected to see an increasing number of transparent AI systems in the market in the near future. In this

*Transparency*: The quality of being open in meaning, in such a way that it is easy for others to see what actions are being performed.

*Opacity:* The quality of being obscure in meaning; also, lacking transparency.

instance, transparency is regarded as a tool to achieve accountability. For example, the COMPAS AI system was challenged in the Craig Loomi criminal case [1]. In his defense, the defendant pointed out that the COMPAS system was gender-biased because it considered his gender as a factor to compute an assessment score. The decision would not have been contested if there was transparency.

## AI Opacity Levels Explained

Figure 1 illustrates the opacity levels of various types of AI systems: neural networks, metaheuristics, machine learning and fuzzy logic. There are differences between algorithms within each category in terms of the level of opacity at different levels.

Neural networks typically receive the input as-is from a data set. The input stage is therefore transparent. Metaheuristics such as the Genetic Algorithm start out with a random initialization of inputs, such as the initial population. Thus, the process is colored opaque. Machine learning systems are typically operating on preprocessed data; already subject to feature engineering or normalization. As the inputs are not entirely raw, the opacity levels are colored grey. Fuzzy logic on the other hand involves opaque modification of the fuzzy inputs based on a certain membership function, and a transparent output. For the operations stage, machine learning includes regression or classification techniques, which are transparent; decision trees and support vector machine operations are transparent, as they are based on explicit mathematical equations. Fuzzy logic, neural networks and metaheuristics are also opaque in the way they update their results through modification or evaluation, before producing an output whereas, machine learning outputs are classes, clusters or coefficients; which fall between fully opaque and fully transparent.

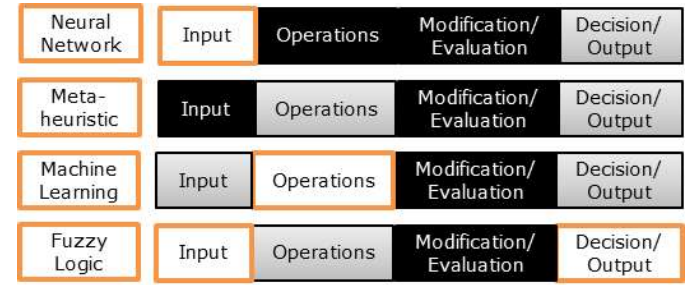

**Figure 1:** Opacity levels of AI algorithms at different stages (light blocks are transparent, grey blocks are semi-transparent/opaque, dark blocks are opaque)

## Explainable AI

Since the concept of AI is often difficult to explain through easily understood principles, the human thoughts related to the domain of human made synthetic knowledge and intelligent decision making capabilities in itself depict a large blackbox with few relatable real-world explanations. As such, various schools of thought exist to categorize AI systems into opaque or transparent [6]. Each of these schools have a different thought pattern to make it easier for humans to mentally grasp the scope and capabilities of AI. By enabling humans to understand and visualize the inner workings of AI systems, they become better equipped to perceive the range of AI capabilities and enable facilitation of magnitude-scale and complex decision-making that the human mind cannot otherwise easily fathom.

The *Logical School* assumes intelligence is synonymous to logic and an AI system is composed of a list of predicates and logical operations. The system is not opaque and though it encompasses complex relationships, it is still interpretable. Similarly, the *Bayesian School* assumes that solutions to any problem can be reached probabilistically. Though randomness exists in the process, the inner functions are expressed through mathematical functions and as such they are not opaque. The *Machine Learning School* is based on statistical concepts and essentially is not opaque for small data sets. However, for improved precision and accuracy, a machine learning system requires large volume data sets, with many dimensions. These two factors make it difficult to interrupt the inner workings of such a system, making them opaque. Deep Neural Networks are extremely opaque, as the underlying mathematics associated with back propagation and weight interactions is very complex and the large number of nodes, layers and interconnecting structures make the system uninterpretable.

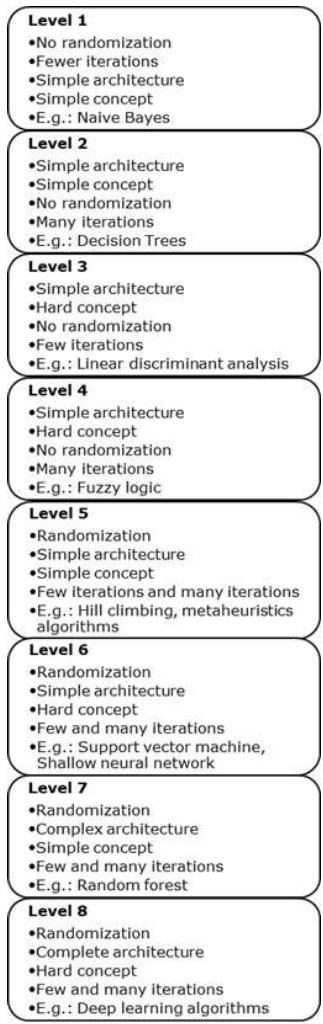

**Figure 2:** Transparency levels of AI systems (most to least transparent – top to bottom)

In order to classify AI systems according to their transparency, we define a taxonomy in Figure 2. The most transparent system (Level 1) is easy to understand, has a simple architecture and runs for a few iterations. In contrast, systems that incorporate randomization and comprise a complex processing architecture based on hard mathematical concepts, regardless of the number of iterations, are considered opaque and they serve as the fundamental definition of a blackbox (Level 8).

## Conclusion

We have identified several categories of AI systems based on their levels of opacity (or transparency). The recent shift towards more transparent AI systems will ensure a better comprehension in terms of the logic and attributions that enable the system to make automated intelligent decisions, guided by human intelligence.

# A Survey of Interpretable AI

Aneesha Bakharia
Institute of Teaching and
Learning Innovation
The University of Queensland
aneesha.bakharia@gmail.com

## Abstract

The question of How can Explainable AI can be achieved? is difficult to address without considering why there exists a need for Explainable AI, the current approaches taken to enable interpretable AI models and the interdisciplinary skills that are required to achieve the ambitious goal of Explainable AI? In order to answer these broad questions, this paper, provides a brief survey of current attempts to provide tools to interpret complex deep and machine learning models from both a mathematically and user interface point of view. Usability and user interface design both play a key role in presenting end-users with a means to make sense of and explore the summary interpretation information. A rationale for taking a multi-disciplinary approach to move beyond simple model interpretation to full explanation is then outlined.

## Author Keywords
Deep Learning, Machine Learning, Interpretable AI, Explainable AI

## ACM Classification Keywords
H.5.m [Information interfaces and presentation (e.g., HCI)]: Miscellaneous.

## Introduction

Some machine learning algorithms are more transparent than others! The decision tree algorithm, as an example, can be expressed as nodes in a tree with criteria used to decide which path in the tree is navigated in order to reach a classification decision. A Support Vector Machine (SVM) on the other hand, separates two classes to be classified by finding a decision boundary that maximizes the separation between the 2 classes. The decisions made by the SVM are mathematically motivated and as a result more difficult to provide the end-user with an interpretable model. Deep learning algorithms which are primarily based on multi-layer neural networks, also do not naturally provide transparent and interpretable results.

Deep learning models however, are currently producing state of the art results on tasks such as vision processing and natural language processing [3]. The high accuracy achieved by deep learning models, is driving the deployment of these models in mission critical domains such as autonomous systems (i.e., self-driving cars) and health care (i.e., health predictive modeling). In these mission critical domains, human life is potentially in danger and the knowledge workers that use these algorithms will need to employ both their own expertise as well as understand the reasons why an algorithm is either directly making decisions or recommending actions. Transparent algorithms are required to establish user trust.

## Current Interpretable AI models

Current research has focused on maintaining the accuracy of deep learning models while at the same time improving a humans ability to understand why decisions have been made. Current research has also been of a mathematical nature and usually undertaken by deep learning algorithm creators. Four recent methods that produce interpretable representations are reviewed in this paper. In this paper interpretable models are viewed as the preceding step to create truly Explainable AI. The definition of interpretable AI follows the definition from [4] which defines interpretability to be "the ability to explain or to present in understandable terms to a human".

Finding out why a decision has been made by a deep learning model is only one aspect of the problem, as there are still questions on how this information should be displayed to the end user (i.e. knowledge worker). Both user interface design and modality of explanation play an important role here.

*Layer-wise Relevance Propagation (LRP)*
LRP [1] works backwards from the probability of a classification, redistributing the probability through preceding neural layers and individual neurons, to the input provided to the neural network. Subsets of inputs of that have achieved high relevance, can then be used to interpret a prediction or classification.

The initial publication describing LRP, uses an image classification example (see Figure 1). In Figure 1, 4 images are shown along with the pixels that have the highest relevance. The pixels with the highest relevance, for these 4 images match the outline of the objects being matched. The end-users ability to interpret the representation depends upon their ability to recognize the objects outline. The level of interpretation will vary across datasets and by the end-users level of domain knowledge. Techniques from user experience design will aid with deciding if the LRP technique will be useful to end-users and inform the user interface.

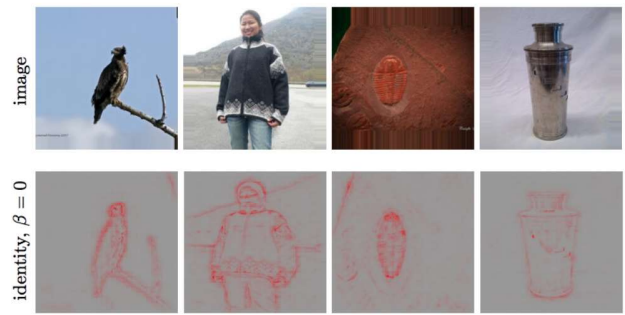

**Figure 1:** Pixels highlighted by the LRP method (reproduced [1])

*Local Interpretable Model-Agnostic Explanations (LIME)*
LIME [5] is perhaps the most well known method to provide an interpretation for classification algorithms that is able to work across both vision and natural language domains. LIME is also algorithm agnostic meaning it can used in conjunction with any algorithm. LIME works by changing the inputs to a model and reviewing how the outputs have changed and builds a decision tree to explain the inner workings of the algorithm. The popularity of LIME is due to the fact that it is open source and provides an intuitive user interface for end-user model interpretation.

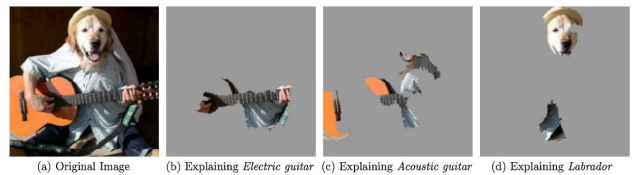

**Figure 2:** Using LIME to interpret image classification (reproduced from [5])

In Figure 2(a) an image of a labrador playing a guitar is shown. Figure 2(b-c) shows the relevant sections of the original image that map to different classifications (i.e., playing a guitar, an acoustic guitar and the labrador). As with the LRP method, the end-users ability to recognise each of the separate objects is required to interpret the predicted classification. As with LRP, the level of interpretation will vary across datasets and by the end-users level of domain knowledge. LIME is not restricted to the visual domain, in the case of text classification, top words that contribute to a decision are highlighted.

*The RETAIN model*
Both LRP and LIME can be applied to static features. In this section, we will briefly review the REverse Time AttentIoN (RETAIN) model [2] which can provide interpretation for time-based sequences. RETAIN uses two recurrent neural networks (to take into consideration sequential information) and an attention mechanism to understand what neurons are firing when a decision is being made. In the initial paper introducing RETAIN, the model was used to understand why certain patients were predicted to be at high risk of heart failure.

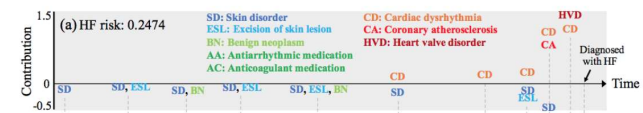

**Figure 3:** Temporal visualization of patents visit records where the contribution of variables for diagnosis of heart failure (HF) is summarized along the x-axis (i.e. time) and the y-axis indicating the magnitude of visit and code specific contributions to the diagnosis (reproduced from [2])

In Figure 3, the output of RETAIN is visualised. The example is very domain specific and would require background medical knowledge to understand the contributing variables. User experience, interface design and visualization techniques could be employed to aid user interpretation and data exploration.

## Discussion

In this paper three (3) methods to aid with the interpretation of deep learning models were briefly introduced. All methods while providing a representation for the basis of interpretation, were only able to highlight low-level features. Evident in all three (3) methods is the need for user experience design, user interface design and visualization design to enhance the ability for an end-user to make sense of the provided information and interpret the algorithms decision. It should also be noted that all three (3) methods, are only able to discover the low-level features (e.g., the outline of an object, the parts of an image or the main words contributing to a classification).

## A multi-disciplinary approach to achieve Explainable AI

It is apparent from the interpretable AI examples presented in this paper, that providing a mathematical underpinning to investigate why a neuron has fired, is not enough to provide interpretable AI. Successful model interpretation, needs to be aided with intuitive user interface design. The interface design needs to display the reasons why decisions have been made in an appropriate representation and the representation may vary depending on the type of domain (e.g. images vs text) and the domain knowledge of the end-user.

Significant contribution from additional discipline areas is required however to achieve Explainable AI. Providing simple user interfaces to review representations may not suffice when complex multi-algorithm systems are used (such as self-driving cars or the move towards artificial general intelligence). These examples in particular require dialogic explanation and human-algorithm argumentation. Discipline contributions from philosophy, psychology and education will be required in addition to algorithm creators and human interaction designers.

# Designing Explainable AI Interfaces through Interaction Design Techniques

Joshua Newn
University of Melbourne
Melbourne, Australia
newnj@unimelb.edu.edu

Eduardo Velloso
University of Melbourne
Melbourne, Australia
evelloso@unimelb.edu.au

Ronal Singh
University of Melbourne
Melbourne, Australia
rr.singh@unimelb.edu.au

Frank Vetere
University of Melbourne
Melbourne, Australia
f.vetere@unimelb.edu.au

Prashan Madumal
University of Melbourne
Melbourne, Australia
pmathugamaba@unimelb.edu.au

## Abstract
The field of Explainable Artificial Intelligence (XAI) aims to design intelligent agents whose predictions, decisions and actions are easily understood by humans upon interrogation. While researchers and practitioners propose a wide variety of solutions, such as building interpretable AI models for generating explanations, we explore the use of human-centred approaches to design the explainable interfaces to communicate explanations to the end-user effectively. In this paper, we present a simple case study of how interaction design can inform the design of an explainable interface for an artificial agent.

## Author Keywords
Explainable AI; Explainable Interfaces; Interaction Design

## Introduction
Imagine spectating a competitive game between two players around a table. You are curious which player will win so you start observing the moves and behaviours of each player from the beginning of the game to predict what each player might do next. A friendly spectator standing beside you asks what you think the next moves might be. Would you be able to offer an accurate prediction? If so, would you be able to explain how you arrived at your prediction? What information and how much information would your explanation contain?

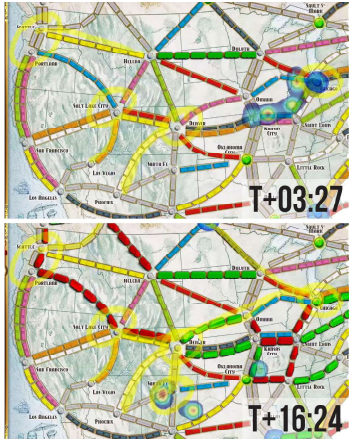

**Figure 1:** In this example, the gaze of the opponent is presented using a real-time heatmap visualisation. At the beginning of the game (top), we can see that the board has little evidence of both players plans. Using gaze we can make early predictions on some cities (circled in yellow), and some inferences on the path they might take (highlighted in yellow). In the later stage of the game (bottom), we see that some of these predictions have materialised. Keeping information hidden is, therefore, core to the game, as a player can gain a significant advantage by correctly guessing their opponents' hidden objectives.

Forming and communicating explanations is no doubt a complex process. Besides knowing how to explain, we also need to select the right level of explanation; an explanation of how something works will fail either if it provides too much detail or if it presupposes too much and skips over essential details [4]. In this work, we focus on two critical aspects of explanation based on the latest XAI literature [3]—*presentation format* and *content*—from an interaction design perspective. Beyond utilising interaction design to improve and evaluate explainable interfaces as suggested by Abdul et al. [1], we wanted to find out if it can be used to elicit natural language explanations. Therefore, we employed the Wizard-of-Oz prototyping technique [6], but asked participants to form appraisals by taking on the role of a predictor-explainer instead of acting as a dialogue-based intelligent agent. We provide details of our study below.

## Study

We designed a study to find out *how* humans formulate predictions and subsequently explain their reasoning process when shown a visual representation of gaze of an opponent in a strategic game. We employ the use of an online multiplayer game called *Ticket to Ride*[1] used in our previous studies for gaze-based intention recognition [7]. Gaze has been shown to be especially beneficial for strategy formulation, especially if a player is able to make accurate and early inferences on the opponent's plans (see Figure 1). We then recruited 20 participants with high proficiency in English (M=25, SD=3.7).

*Experimental Setup*
To simplify the study, we used screen recordings from the perspective of a 'gaze aware' player who was given the ability to see a visual overlay of their opponent's gaze while playing the digital version of *Ticket to Ride* game. This avoided us from recruiting pairs of players for this part of the study. The visual overlay was in the form of a real-time dynamic heatmap visualisation. Each recording was of an actual game played, and we used each recording only once as we wanted to elicit a wide range of textual representations from different game scenarios. As we adopt a Wizard-of-Oz approach for our setup, we employed a protocol alongside to continually reinforce the participants' belief that they were engaged in a live online game with two other players for the duration of the study. The researcher was allowed to clarify the rules about the game when prompted, but that was the extent of their interference.

The technical setup consisted of a computer connected to two 23-inch monitors with two pairs of keyboards/mice on a rectangular table in a lounge room setting. The participant and experimenter sat at opposite ends of the table so that the experimenter's display was not visible to the participant.

*Procedure*
Upon arrival, participants were asked to be seated and to provide their consent. We then informed that they would assist one of the players in a 2-player game taking place in another room for one game. We informed participants that they had been 'randomly' selected to take on the role of an adviser and that their 'teammate' would be the player receiving assistance from them while they played the game. As each session is designed to last a maximum of an hour, we informed the participant in advance that the game would begin at a fixed time, typically partway through the session (e.g. 11:30) to reinforce the deception. We then mention that their teammate has to play the game's tutorial while they got familiarised with being an adviser. All participants went through a familiarisation step until it was 'time to join the game' despite their experience with the game to get participants to start thinking about forming *predictions* and *explanations* from intentions derived from gaze observation as it was an unfamiliar task to many. The phase also potentially instil confidence in the participant when observ-ing and communicating at the same time in the upcoming 'live' game. We reminded participants to provide messages that their teammate would find helpful and to build the teammate's trust by being transparent in how they derive their predictions through their explanations. Next, we demonstrated the simple chat application that served as the means of communication with their non-existent teammate (see Figure 2).

---
[1] http://www.daysofwonder.com/tickettoride/en/

**Figure 2:** The chat application contains two text fields for their *prediction* and *explanation* respectively, a send button and a window showing the conversation. The application logs all messages sent and included a validation to ensure both text fields are filled.

We posed no restrictions on the language format participants could use for communication, allowing them to freely formulate their messages as long as each message contained a prediction of their opponent's intentions followed by an explanation. At the end of the study, the participant was given a short questionnaire on their experience, followed by a brief interview based on their responses and strategies employed.

## Results

The ability to successfully formulate messages depended on several factors, including individual ability, experience with the game, the communication strategy adopted and the recorded game shown itself. We obtained a total of 246 prediction-explanation messages, ranging from abstract to detailed.

### Prediction Format

For each prediction part of a message, we stripped them into its essential and meaningful components to obtain a minimal format for prediction, which gave us a total of 45 initial formats. We merged formats that were similar in nature into 11 key prediction formats (see Table 1). We noted participants conveyed their level of confidence when providing their predictions, using phrases that express uncertainty (e.g. i think/maybe/will try).

### Explanation Content

Participants provided a wide range of explanations for their predictions. For the purposes of this work, we are interested in complex explanations as opposed to simple explanations. We found that complex explanations contain *spatial*, *temporal* and

**Table 1:** Prediction Formats

| Pattern | N |
| --- | --- |
| from [City/Area] to [City/Area] | 130 |
| from [City/Area] to [City/Area] through $n$[City] | 37 |
| to/towards [City/Area] | 35 |
| interest in/along/around [Area] | 15 |
| [Intentional Action] | 8 |
| around [City] | 5 |
| to [City/Area] through [City/Area] | 4 |
| between [City] and [City] | 4 |
| from [City/Area] | 3 |
| from/to [City] to/from [City] or [City] | 3 |
| from [City] to [City] in/to [Area] | 2 |

*quantitative* properties, in line with Dodge et al. [2]'s findings on expert explanations. In order to build a general model, we turn to [5]'s explanation model for labelling the properties for complex explanations with the assumption that the model can be generalised to explain nonverbal input such as gaze.

Explanations can be derived from different sources of information available to the agent. This includes information about past and potential future actions derived in accordance with [5]'model. This involves *Causal History of Reasons*, defined as $O_a$, and *Intentional Action*, defined as $I_a$. Our results show that participants had a strong reliance on gaze to explain the predictions. We believe that gaze being 'always on', its information became more prominent throughout the game for enabling predictions as compared to observable actions. For this reason, we include gaze ($O_g$) as part of every explanation generated using our piece-wise function below.

$$Explanation = \begin{cases} O_g, O_a & \text{if ontic actions observed} \\ O_g, I_a & \text{if intentional action likely} \\ O_g, I_a, O_a & \text{otherwise} \end{cases} \quad (1)$$

Therefore, combination of all three sources of information forms an explanation that is highly detailed, for example:

> "The opponent is building a route from Washington to New Orleans through Nashville in the South East [*Prediction (i)*]. The opponent has claimed part of this route [$O_a$], has been looking at the routes between Raleigh and Little Rock repeatedly [$O_g$] and is likely to claim Nashville to Raleigh next [$I_a$]."

*Communication Strategies*
Participants adopted two general strategies for communicating the intention of their opponent. The first strategy was to observe and provide the best possible prediction accompanied by a detailed explanation. This strategy resulted in fewer predictions; especially if the current prediction or reasoning does not change. The second type of strategy employed was to provide as many predictions as possible which they noted was limited to how fast they could type. As participants formed these messages, they noted they had a fear of missing out on observations that might be important.

## Discussion & Conclusion
Our study presents a simple case of how interaction design techniques can be used to inform the design of explainable interfaces. Our study resulted in the basis for a computational model of explanation, in which we can use gaze and ontic actions to form explanations, and we can vary the level of detail as needed. Beyond answering the *how* and *what* questions to meet our design goals, we learn it is important to know *when* to provide an explanation in the context of predictions, and this requires the agent to be contextually-aware of the what the assisted-player already knows and whether the information is helpful to them. We also learn that it is possible to portray uncertainty when communicating predictions in a natural language, providing an alternative to using confidence levels. Our next step is to evaluate the language model built using an AI-assisted player using interaction design.

# A Grounded Dialog Model for Explainable AI

**Prashan Madumal**
University of Melbourne
Melbourne, Australia
pmathugamaba@unimelb.edu.au

**Tim Miller**
University of Melbourne
Melbourne, Australia
tmiller@unimelb.edu.au

**Frank Vetere**
University of Melbourne
Melbourne, Australia
f.vetere@unimelb.edu.au

**Liz Sonenberg**
University of Melbourne
Melbourne, Australia
l.sonenberg@unimelb.edu.au

## Abstract
To generate trust with their users, Explainable Artificial Intelligence (XAI) systems need to include an explanation model that can communicate the internal decisions, behaviours and actions to the interacting humans. Successful explanation involves both cognitive and social processes. In this paper we focus on the challenge of meaningful interaction between an explainer and an explainee and investigate the structural aspects of an interactive explanation to propose a explanation dialog model.

## Author Keywords
Explainable AI; Dialog models

## Introduction
Artificial intelligence (AI) systems that aim to be transparent about their decisions must have understandable explanations that clearly justify their decisions. This is especially true in scenarios where people are required to make critical decisions based on the outcomes of an AI system. An appropriate explanation can promote trust in the system, allowing better human-AI cooperation [11]. Explanations also help people to reason about the extent to which, if at all, they should trust the provider of the explanation.

However, much research and practice in explainable AI uses the researchers' intuitions of what constitutes a 'good'

explanation rather basing the approach on a strong understanding of how people define, generate, select, evaluate, and present explanations [9, 10]. Most modern work on Explainable AI, such as in autonomous agents [14, 2, 3, 6] and interpretable machine learning [5], does not discuss the interaction and the social aspect of the explanations. The lack of a general interaction model of explanation that takes into account the end user can be attributed as one of the shortcomings of existing explainable AI systems. Although there are existing conceptual explanation dialog models that try to emulate the structure and sequence of a natural explanation [1, 12], we propose that improvements will come from further empirically-driven study of explanation.

Understanding how humans engage in conversational explanation is a prerequisite to building an explanation model, as noted by Hilton [8]. De Graaf [4] note that humans attribute human traits, such as beliefs, desires, and intentions, to intelligent agents, and it is thus a small step to assume that people will seek to explain agent behaviour using human frameworks of explanation. We hypothesise that AI explanation models with designs that are influenced by human explanation models have the potential to provide more intuitive explanations to humans and therefore be more likely to be understood and accepted. We suggest it is easier for the AI to emulate human explanations rather than expecting humans to adapt to a novel and unfamiliar explanation model. While there are mature existing models for explanation dialogs [12, 13], these are idealised conceptual models that are not grounded on or validated by data, and seem to lack iterative features like cyclic dialogs.

In this paper our goal is to introduce an dialog model and interaction protocol that is based on data obtained from different types of explanations in actual conversations.

## Data

We collected data from six different data sources encompassing six different types of explanation dialogs. Table 1 shows the explanation dialog types, explanation dialogs that are in each type and number of transcripts.

We formulate our design based on an inductive approach. We use grounded theory [7] as the methodology to conceptualize and derive models of explanation. The key goal of using grounded theory, as opposed to using a hypothetico-deductive approach, is to formalize a model that is grounded on actual conversation data of various types, rather than a purely conceptual model.

Table 1: Coded data description.

| Explanation Dialog Type | #Dialogs | #Scripts |
| --- | --- | --- |
| 1. Human-Human static explainee | 88 | 2 |
| 2. Human-Human static explainer | 30 | 3 |
| 3. Human-Explainer agent | 68 | 4 |
| 4. Human-Explainee agent | 17 | 1 |
| 5. Human-Human QnA | 50 | 5 |
| 6. Human-Human multiple explainee | 145 | 5 |

## Explanation Dialog Model

We present our explanation dialog model derived from the study. Figure 1 depicts the derived explanation dialog model as a state diagram where the sequence of each component is preserved. The labels 'Q' and 'E' refer to the Questioner (the explainee) and the Explainer respectively.

We identify two loosely coupled sections of the model: the *Explanation Dialog* and the *Argumentation Dialog*. These two sections can occur in any order, frequencies and cycles.

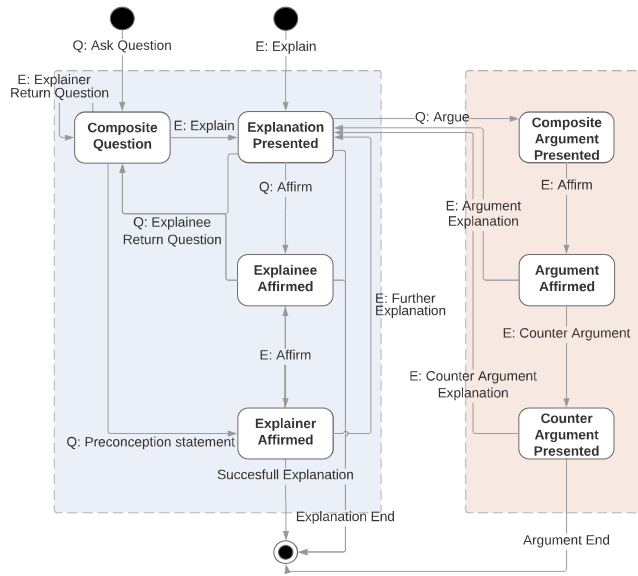

**Figure 1:** Explanation Dialog Model

A typical example would be, an explainee asking a question, receiving a reply, presenting and argument of the explanation, explainer acknowledging the argument, explainer agreeing to the argument. This scenario can be described in states as Start →Composite Question →Explanation Presented →Explainee Affirmed →Argumentation Dialog Initiated →Composite Argument Presented →Argument Affirmed →End.

## Analysis

We focus our analysis on three areas to further reinforce the derived interaction protocol: 1. Key components of an Explanation Dialog; 2. Relationships between these components and their variations between different dialog types; and 3. The sequence of components that can successfully carry out an explanation dialog.

*Code Frequency Analysis*

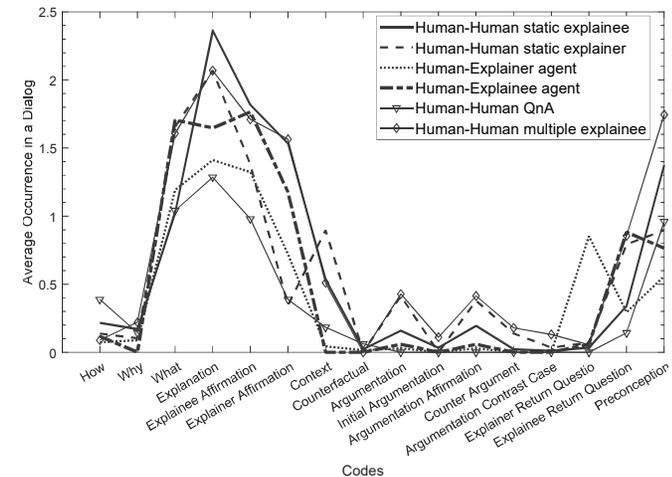

**Figure 2:** Average code occurrence per dialog in different explanation dialog types

The average code occurrence per dialog in different dialog types is depicted in Figure 2. In all dialog types, a dialog is most likely to have multiple *what* questions, multiple *explanations* and multiple *affirmations*.

## Conclusion

Explainable Artificial Intelligent systems can benefit from having a proper interaction protocol that can explain their actions and behaviours to the interacting users. In this paper, we propose a dialog model for the socio-cognitive process of explanation that is derived from different types of natural conversations between humans as well as humans and agents.